\documentclass[a4paper,twoside]{article}

\usepackage{epsfig}
\usepackage{subcaption}
\usepackage{calc}
\usepackage{amssymb}
\usepackage{amstext}
\usepackage{amsmath}
\usepackage{amsthm}
\usepackage{multicol}
\usepackage{pslatex}
\usepackage{apalike}
\usepackage{algorithm2e}
\usepackage[bottom]{footmisc}
\usepackage{hyperref}

\usepackage{tabularx}
\usepackage{rotating}
\usepackage{float}
\usepackage{booktabs}
\usepackage{multirow}
\usepackage{bbold}
\newcolumntype{Y}{>{\centering\arraybackslash}X}
\usepackage{SCITEPRESS}     

\begin{document}
\title{Improving Pseudo-labelling and Enhancing Robustness for Semi-Supervised Domain Generalization}
\author{\authorname{Adnan Khan\sup{1}, Mai A. Shaaban\sup{1} and Muhammad Haris Khan\sup{1}}
\affiliation{\sup{1}Mohamed bin Zayed University of Artificial Intelligence, Abu Dhabi, U.A.E}
\email{\{adnan.khan, mai.kassem, muhammad.haris\}@mbzuai.ac.ae}
}
\abstract{Beyond attaining domain generalization (DG), visual recognition models should also be data-efficient during learning by leveraging limited labels. We study the problem of Semi-Supervised Domain Generalization (SSDG) which is crucial for real-world applications like automated healthcare. SSDG requires learning a cross-domain generalizable model when the given training data is only partially labelled. Empirical investigations reveal that the DG methods tend to underperform in SSDG settings, likely because they are unable to exploit the unlabelled data. Semi-supervised learning (SSL) shows improved but still inferior results compared to fully-supervised learning. A key challenge, faced by the best-performing SSL-based SSDG methods, is selecting accurate pseudo-labels under multiple domain shifts and reducing overfitting to source domains under limited labels. In this work, we propose new SSDG approach, which utilizes a novel uncertainty-guided pseudo-labelling with model averaging (UPLM). Our uncertainty-guided pseudo-labelling (UPL) uses model uncertainty to improve pseudo-labelling selection, addressing poor model calibration under multi-source unlabelled data. The UPL technique, enhanced by our novel model averaging (MA) strategy, mitigates overfitting to source domains with limited labels. Extensive experiments on key representative DG datasets suggest that our method demonstrates effectiveness against existing methods. Our code and chosen labelled data seeds are available on GitHub: \href{https://github.com/Adnan-Khan7/UPLM}{https://github.com/Adnan-Khan7/UPLM} .}
\onecolumn \maketitle \normalsize \setcounter{footnote}{0} \vfill
\section{\uppercase{Introduction}}
\label{sec:intro}
\begin{figure*}[!htbp]
  \vspace{-0.2cm}
  \centering
 {\epsfig{file = 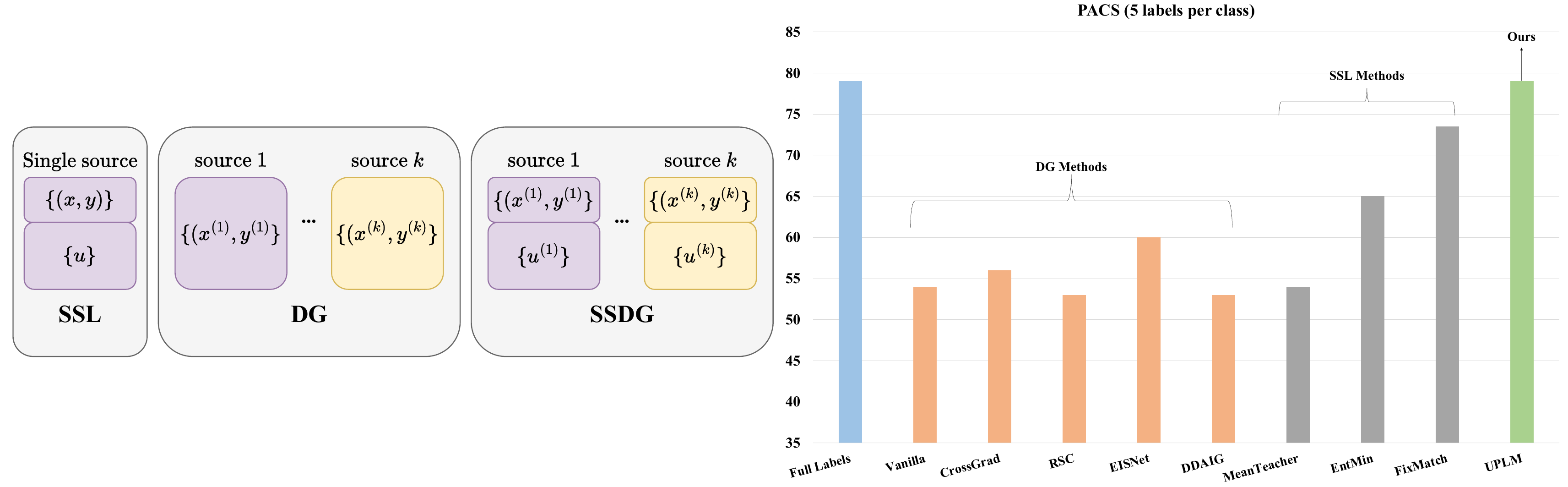, width = 15.5cm}}
\caption{(left) Visual comparison of SSL, DG, and SSDG setting. (right) Performance comparison of three paradigms.}
\label{fig:comparison of dg and ssdg}
\end{figure*}

Domain shift \cite{tzeng2015simultaneous} \cite{hoffman2017simultaneous} is an important challenge for several computer vision tasks e.g., object recognition \cite{krizhevsky2017imagenet}. Among others, domain generalization (DG) has emerged as a relatively practical paradigm for handling domain shifts and it has received increasing attention in the recent past \cite{li2017deeper}  \cite{zhou2021domain} \cite{khan2021mode}. The goal is to train a model from the data available from multiple source domains that can generalize well to an unseen target domain. We have seen several DG approaches \cite{huang2020self} \cite{wang2020learning}  that have displayed promising performance across various benchmarks \cite{li2017deeper} \cite{venkateswara2017deep}.  
However, the performance of many DG methods is sensitive to the availability of sufficiently annotated quality data from available source domains. As such, this requirement is difficult to meet in several real-world applications of these models e.g., healthcare, autonomous driving and satellite imagery \cite{10121898}. Besides attaining generalization, it is desirable for the learning algorithms to be efficient in their use of data. This means that the model can be trained using a minimal amount of labelled data to reduce development costs. This concept is closely related to semi-supervised learning (SSL) \cite{grandvalet2004semi} \cite{tarvainen2017mean}  which seeks to make use of large amounts of unlabelled data along with a limited amount of labelled data for model training.
To this end, this paper studies the relatively unexplored problem of semi-supervised domain generalization (SSDG). It aims to tackle both the challenges of model generalization as well as data-efficiency within a unified framework. Both DG and SSDG share the common goal of training models capable of performing well on unseen target domain using only source domain data for training. However, DG is based on the assumption that all data from source domains is fully labelled, while SSDG operates under the SSL setting, where only few images within each source domain have labels and large number of images are unlabelled. Figure~\ref{fig:comparison of dg and ssdg} shows the visual comparison among the settings of three related paradigms.

We note that, the DG methods, which cannot utilize unlabelled data, tend to show degraded performance upon reducing the quantity of labelled data \cite{zhou2021semi}. On the other hand, the SSL methods e.g, FixMatch \cite{sohn2020fixmatch} display relatively better performance than the DG methods under limited labels setting, still their performance is noticeably inferior to the fully labelled setting. SSL methods lose performance in SSDG setting due to differences in data distributions between various source domains and the limited amount of labelled data available, which are unique challenges to SSDG problem.

We propose a systematic approach, namely \textbf{U}ncertainty-Guided \textbf{P}seudo-\textbf{L}abelling with \textbf{M}odel Averaging (UPLM) to tackle the challenges in SSDG. First, we develop an uncertainity-guided pseudo-labelling (UPL) technique to overcome the problem of noisy pseudo-labels (PLs), typically produced by confidence-based methods \cite{sohn2020fixmatch} under domain shift. We leverage model's predictive uncertainty to develop a pseudo-label selection criterion that provides accurate PLs by mitigating the impact of miscalibrated predictions, especially for out-domain data. Second, we propose a novel model averaging (MA) technique which  overcomes the effect of overfitting to limited labels in source domains to achieve cross-domain generalization at the inference stage. Through empirical results we show the intuition and motivation behind our two components. Our suggested approach demonstrates its effectiveness in addressing the SSDG problem when compared to other SSDG and SSL methods, as evidenced by thorough experimentation on four demanding DG datasets.

\section{\uppercase{Related Work}}
\label{sec:related_work}
\noindent\textbf{Domain generalization.} \cite{vapnik1999nature} is recognized as pioneering work in Domain Generalization (DG), introducing Empirical Risk Minimization (ERM) to minimize the sum of squared errors across diverse source domains. It led to various approaches for extracting domain-invariant features, such as \cite{muandet2013domain} employing maximum mean discrepancy (MMD), \cite{ghifary2015domain} introducing a multi-task autoencoder, and \cite{yang2013multi} using canonical correlation analysis (CCA). Meta-learning frameworks, like those in \cite{shu2021open}, have also been employed for domain generalization to simulate training domain shifts.
For semantic alignment, domain generalization such as \cite{kim2021selfreg} and \cite{dou2019domain}, leverage self-supervised contrastive formulations \cite{khan2022contrastive}. The idea of improving diversity in source domains is shown to be effective for DG \cite{khan2021mode}. \cite{volpi2018generalizing} applied a wasserstein constraint in semantic space and \cite{shankar2018generalizing} introduced Crossgrad training as a DG method to enhance DG.
The aforementioned DG methods assume supervised settings with fully labelled source domain data for training. However, there is limited research on enhancing DG performance in scenarios with scarce labelled data. This work addresses the SSDG problem, unifying data efficiency and model generalization, and proposes a principled approach to tackle relevant SSDG challenges.

\noindent\textbf{Uncertainty estimation in DNNs.} Quantifying uncertainty in deep nerual networks (DNNs) has remained an important research direction \cite{kendall2016modelling}. These methods proposed to quantify the uncertainty associated with the predictions made by DNNs. For instance, \cite{gal2016dropout} presented dropout training in DNNs as approximate Bayesian inference to model uncertainty. \cite{kendall2017uncertainties} developed a Bayesian DNNs framework that combines input-dependent aleatoric uncertainty with epistemic uncertainty. The work of \cite{lakshminarayanan2017simple} proposed alternative to Bayesian DNNs, which includes ensembles and adversarial training, for estimating predictive uncertainty on out-of-distribution examples. \cite{smith2018understanding} investigated measures of uncertainty focusing on mutual information and proposed an improvement in uncertainty estimation using probabilistic model ensembles. In this work, we leverage model uncertainty from Monte-Carlo (MC) dropout technique \cite{gal2016dropout} which is used to develop a pseudo-label selection criterion under multiple domain shifts in the SSDG problem.

\noindent\textbf{Semi-supervised domain generalization.} The problem setting in DG assumes fully-supervised settings i.e., the source domains data is completely labelled. In many real-world deployment scenarios, however, this is a strict requirement, as it is costly and some times infeasible to acquire sufficiently labelled data. To address this limitation, a more practical and widely applicable setting is semi-supervised domain generalization (SSDG), which combines model generalization and data efficiency into a single paradigm. 
For instance, \cite{lin2021semi} introduced a cyclic learning framework to enhance model generalization by promoting a positive feedback between the pseudo-labelling and generalization phases. The authors in \cite{zhou2021semi} proposed StyleMatch as an effective approach that extends FixMatch with stochastic modeling and multi-view consistency learning to achieve significant improvements in SSDG problem. \cite{yao2022enhancing} proposed confidence-aware cross pseudo supervision algorithm that utilizes Fourier transformation and image augmentation to enhance the quality of PLs for SSDG medical image segmentation. \cite{qi2022multimatch} proposed MultiMatch, a SSDG method that extends FixMatch to the multi-task learning framework, utilizing the theory of multi-domain learning to produce high-quality PLs.

\section{\uppercase{Proposed Framework}}
\label{sec:proposed_framework}
The problem of semi-supervised domain generalization (SSDG) has two distinct challenges: (1) how to obtain accurate pseudo-labels under multiple domain shifts, and (2) reduce overfitting to source domains under limited labels. To this end, we present a principled approach to SSDG namely uncertainty-guided pseudo-labelling (Section~\ref{subsection:Uncertainty-guided_Pseudo-Labelling}) with model averaging (Section~\ref{subsection:Model Averaging}) to counter the two challenges in SSDG. 

\noindent \textbf{Problem Settings.} We first define few notations and then present the formal definition of SSDG. Formally, $\mathcal{X}$ and $\mathcal{Y}$ denote the input and label spaces, respectively. A domain is a combination of the (joint) probability distributions for $X$ and $Y$, denoted by $P(X,Y)$, over the corresponding spaces $\mathcal{X}$ and $\mathcal{Y}$. We use $P(X)$ and $P(Y)$ to show the marginal distributions of $X$ and $Y$, respectively. Our focus in this study is on distribution shifts only in $P(X)$, while $P(Y)$ remains constant. This means that all domains share the same label space.
Similar to DG, in SSDG, we are provided with $K$ distinct but related source domains $\mathcal{D} = \{\mathcal{D}\}_{k=1}^K$, where $\mathcal{D}_k$ denotes the distribution over the input space $\mathcal{X}$ for domain $k$, and $K$ is the total number of source domains. 
From each source domain $\mathcal{D}_k$, we are provided with a labelled set comprising of input-label pairs $\mathcal{D}_k^L=\{(x^{k},y^{k})\}$ and an unlabelled set $\mathcal{D}_k^U=\{u^{k}\}$. Note that, $|\mathcal{D}_k^U| \gg |\mathcal{D}_k^L|$.
We also assume the existence of a set of target domains $\mathcal{T}$ typically set to 1. The objective in SSDG is to leverage the labelled set $\mathcal{D}_k^L$ from the source domains, along with the unlabelled data $\mathcal{D}_k^U$, to learn a mapping $\mathcal{F}_\theta : \mathcal{D}_k^L \cup \mathcal{D}_k^U \rightarrow \mathcal{Y}$ that can provide accurate predictions on data from an unseen target domain $\mathcal{T}$.

\noindent \textbf{Semi-supervised DG pipeline.} We instantiate our proposed method in FixMatch, which is an SSL method and performs better than all DG methods in SSDG settings (Figure~\ref{fig:comparison of dg and ssdg}). It combines consistency regularization \cite{sajjadi2016regularization} and pseudo-labelling (PL) \cite{xie2020self} techniques to achieve state-of-the-art results on several SSL benchmarks. The algorithm consists of two standard cross entropy losses, the supervised loss $\mathcal{L}_{s}$, and an unlabelled loss $\mathcal{L}_u $. The supervised loss is calculated as: $ \mathcal{L}_{s} =
-\frac{1}{|\mathcal{S}|}\sum_{j \in \mathcal{S}} y_{j} \log(\hat{y}_{j})$, 
where $\mathcal{S} = \mathcal{D}_{k=1:K}^S$ is the aggregation of labelled set from all $K$ source domains. $y_{j}$ and $\hat{y}_{j}$ is the ground truth and the predicted probability for $j^{th}$ labelled example, respectively. 

Two augmented versions of the an unlabelled example $u$ are generated i.e., weak and strong augmentations \cite{devries2017improved} denoted by $u'$ and $u''$, respectively. Let $q_{u'}$ and $q_{u''}$ be the predicted probability distributions for $u'$ and $u''$, respectively. For a weakly augmented unlabelled example $u'$, the pseudo-label $\tilde{y}_u{'}$ is generated if $g_{u'}$ is 1, where $g_{u'}$ is a binary variable and obtained as follows: $g_{u'} = \mathbb{1} \left[ \max \left(q_{u'}\right) \geq \tau \right]$, $\tau$ is a scalar hyperparameter denoting the confidence threshold.
The cross entropy (CE) loss is used at the model output for a strongly augmented version $u''$, which introduces the form of consistency regularization to reduce the discrepancy between $u'$ and $u''$. The unsupervised loss $\mathcal{L}_{u}$ becomes: $\mathcal{L}_{u}=
\frac{1}{|\mathcal{U}|}\sum_{u \in \mathcal{U}} \mathbb{1} \left[ \max \left(q_{u'}\right) \geq \tau \right] \mathrm{CE}(\tilde{y}_u{'},q_{u''})$, where $\mathcal{U} = \mathcal{D}_{k=1:K}^U$ is the aggregated unlabelled set from all $K$ source domains. The overall loss then becomes: $\mathcal{L}_{final} = \mathcal{L}_s + \lambda \mathcal{L}_u$, where $\lambda$ is the weight given to the unsupervised loss.
The presence of unlabelled data in different source domains, manifesting different shifts, pose a challenge to confidence-based selection of PLs leading to generation of noisy PLs. Due to various domain shifts, the model is more prone to generating a high confidence for an incorrect prediction, which will then translate into a noisy pseudo-label. To this end, we leverage model's predictive uncertainty to develop pseudo-label selection criterion which allows countering the poor calibration of model, leading to the selection of accurate PLs.

\subsection{Uncertainty-guided Pseudo-Labelling (UPL)}
\label{subsection:Uncertainty-guided_Pseudo-Labelling}

We describe our uncertainty-guided pseudo-labelling (UPL) mechanism to address the challenge of noisy PLs when the unlabelled data could be from different (source) domains. We first quantify the model's predictive uncertainty and then leverage it to construct a uncertainty-guided pseudo-label selection criterion. 

\noindent\textbf{Uncertainty quantification.} We choose to use the Monte-Carlo (MC) dropout method to quantify model's predictive output uncertainty $\mathcal{V}_{u'}$ for an unlabelled example $u'$. It requires the addition of a single dropout layer $(\mathcal{D})$ that is incorporated between the feature extractor network and the classifier. The MC dropout technique requires $\mathcal{N}$ Monte-Carlo forward passes for an unlabelled example $u'$ through the model.
This produces a distribution of probability outputs denoted as $\mathbf{c}_{u'} \in \mathbb{R}^{\mathcal{N} \times C}$ where $C$ is the number of classes. Now, we obtain the uncertainty $\mathcal{V}_{u'} \in \mathbb{R}^{C}$ by computing the variance along the first dimension of $\mathbf{c}_{u'}$. Finally, the $\mathcal{V}_{u'}$ is transformed using $\tanh$ function, to obtain a measure of model certainty $\kappa_{u'}$: $\kappa_{u'} = (1 - \tanh(\mathcal{V}_{u'}))$. 

\begin{figure*}[!htbp]
  \vspace{-0.2cm}
  \centering
 {\epsfig{file = 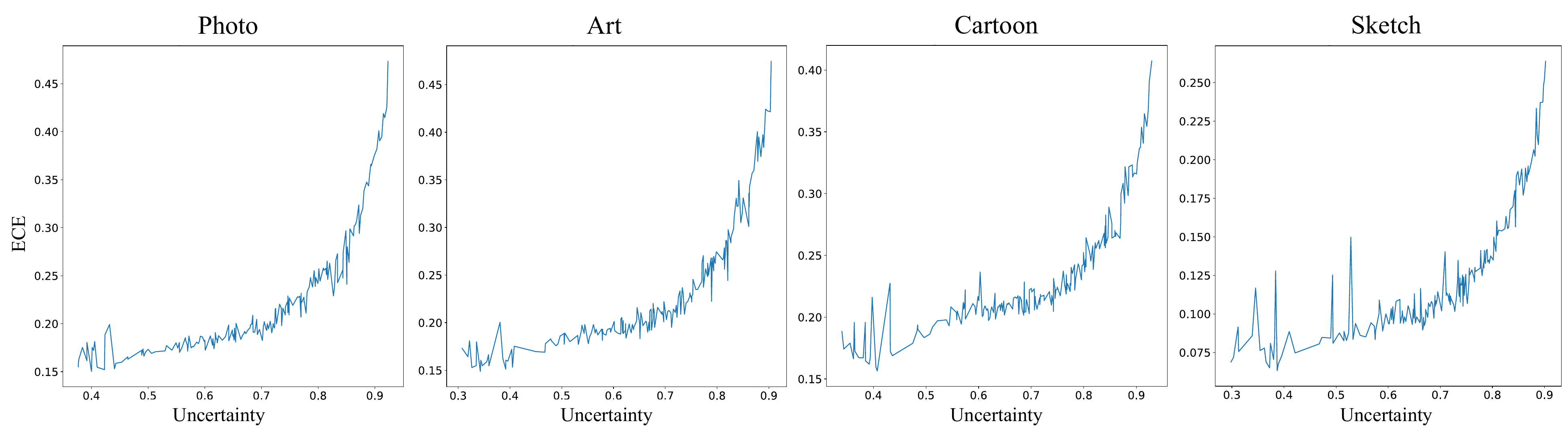, width = 15cm}}
\caption{Uncertainty of selected PLs vs  Expected Calibration Error (ECE).}
\label{fig:ECE} 
\end{figure*}

\noindent\textbf{Uncertainty constraint in PL selection.} In SSDG, due to domain shifts, for an unlabelled input, a model can yield high confidence for an incorrect prediction. This happens because the model is typically poorly calibrated for out-domain predictions. So a confidence-based PL selection criterion is prone to generating noisy PLs. To implicitly mitigate the impact of poor calibration of the model under various domain shifts, motivated by \cite{rizvedefense}, we develop a pseudo-label selection criterion that uses both the predictive confidence and predictive uncertainty of a model.
Specifically, for an unlabelled weakly augmented example $u'$, given the confidence of the predicted class label as: $\max \left(q_{u'}\right)$ and the corresponding certainty as $\kappa_{u'}(\arg\max \left(q_{u'}\right))$. The $\max \left(q_{u'}\right)$ should be greater than the confidence threshold $\tau$ and at the same time the $\kappa_{u'}(\arg\max \left(q_{u'}\right))$ should be greater than the certainty threshold $\eta$ as:  
\begin{equation}
g_{u'} = \mathbb{1} \left[ \max \left(q_{u'}\right) \geq \tau \right]    \mathbb{1}\left[ \kappa_{u'}(\arg\max \left(q_{u'}\right)) \geq \eta \right]
\label{eq:upl}
\end{equation}
In Figure~\ref{fig:ECE}, we plot the relationship between the model output uncertainty and its Expected Calibration Error (ECE) (see also Appendix~\ref{appendix:ece}). It shows that in all cases when the uncertainty of selected PLs increases, the ECE increases and vice versa. Therefore, choosing PLs that are both certain and confident will likely lead to better PL accuracy via counteracting the negative effects of poor calibration.
\subsection{Model Averaging (MA)}
\label{subsection:Model Averaging}
In the training stage, the model may overfit to the limited labelled (or pseudo-labelled) data and eventually perform poorly on unseen target domain data. This problem exacerbates when we introduce hard constraints on PLs selection. Consequently the robustness of a model against domains shifts gets affected and could lead to convergence at poor optimum. To address this, we propose a simple yet effective model averaging (MA) technique at the inference stage. Specifically, we take the weighted average of the model parameters obtained from the best performing model on on held-out validation set ($\theta_{best}$), the model checkpoint from last epoch ($\theta_{last}$), and the exponential moving average model ($\theta_{ema}$). 
The predictions of the three models are averaged using the combined state dictionary, which is created by taking the average of the corresponding weights of the three models denoted by $\theta_{avg}$ given as:
    $\theta_{avg} = \alpha \cdot \theta_{best} + \beta \cdot \theta_{last} + \gamma \cdot \theta_{ema}$
where $\alpha$, $\beta$, and $\gamma$ are the weights assigned to each model. We set $\alpha$, $\beta$, and $\gamma$ to $1/3$ each, indicating that we give equal importance to each model. $\theta_{avg}$ is then used to make predictions on the test data. By using $\theta_{avg}$ model, we reduce the reliance on a single model and its parameters, which leads to better generalization at inference stage.
\section{EXPERIMENTS}
\label{sec:experiments}
\noindent \textbf{Datasets.} We evaluate on four distinct DG datasets: PACS \cite{li2017deeper} (9,991 images, 7 classes, four domains), OfficeHome \cite{venkateswara2017deep} (15,588 images, 65 classes, four domains), TerraIncognita \cite{beery2018recognition} (24,778 images, 10 classes, four domains), and VLCS \cite{fang2013unbiased} (10,729 images, 5 classes, four domains).

\noindent \textbf{Training and implementation details.}
We follow the evaluation protocol of \cite{gulrajani2020search}. For model selection we use the training domain validation protocol. We partition the data from each training domain in 90\% training and 10\% validation subsets and use only 10 labels per class from each source domain. The model that maximizes the accuracy on validation set is considered the best model which is then evaluated on the target domain to report classification (top-1) accuracy. All experiments use an NVIDIA Quadro RTX 6000 GPU with 24GB dedicated memory. We use ResNet-50 \cite{resnet} model as a backbone with a batch size $B$ of 24  for labelled data and $\mu\times B$ for unlabelled data where $\mu = 5$. We use the SGD \cite{robbins1951stochastic} optimizer and train model for 20 epochs (512 iterations each). The learning rate is set to 0.03 with nesterov momentum \cite{nesterov1983method}. We do grid search in the range \{0.2,0.9\} using the validation set for the hyperparameter $\eta$ in UPL method. The optimal $\eta$ values are 0.2, 0.5, 0.5 and 0.7 for PACS, TerraIncognita, OfficeHome and VLCS respectively. We report accuracy for target domains and their average, where a model is trained on source domains and evaluated on an (unseen) target domain. Each accuracy on the target domain is an average over three different trials with different labelled examples. Appendix \ref{appendix:hyperparameters} shows the ablation on different hyperparameters including the number of MC forward passes $\mathcal{N}$, certainty threshold $\kappa$ and the parameter $\mu$ which governs the proportion of unlabelled data within each training batch.

\subsection{Results}
\label{subsection:results}
We investigate the impact of each proposed component for all four datasets. Table~\ref{tab:results_std} presents a comparison of test accuracies achieved by four different methods: FixMatch (baseline), uncertainty-guided PL approach (UPL), model averaging (MA), and our final model (UPLM) across different target domains of four benchmark datasets.
The average test accuracy across all target domains is also shown for each dataset. 
The results demonstrate that the UPLM method achieves the highest test accuracy in three out of four datasets, with an average test accuracy of 78.94\% for PACS, 50.61\% for OfficeHome, 62.72\% for VLCS, and 30.19\% for TerraIncognita. On OfficeHome the constraint of uncertainty limits the number of PLs, and hence relatively less improvement is seen in UPLM as compared to MA. For instance, the class to training examples ratio for VLCS and OfficeHome is 1:2146 and 1:238 respectively. Enforcing an uncertainty constraint on this small set of examples reduces their number even further, making the ResNet-50 model more prone to overfitting. Overall, the UPLM method outperforms in the most target domains across all datasets, indicating that the uncertainty-guided PL approach with model averaging leads to improved performance in SSDG.
\begin{table}[htbp!]
	\centering
	\caption{Comparison of FixMatch, UPL, MA, and UPLM.}
	\scalebox{0.65}{\begin{tabular}{|ll|l|l|l|l|}
		\hline
		&
		\multicolumn{1}{l|}{Target} &
		\multicolumn{1}{c|}{FixMatch} &
		\multicolumn{1}{c|}{UPL} &
		\multicolumn{1}{c|}{MA} &
		\multicolumn{1}{c|}{UPLM (Ours)} \\ \hline\hline
		\multirow{5}{*}{\begin{turn}{90}PACS\end{turn}}
		& Photo           & $82.67_{\pm4.73}$& $89.76_{\pm3.12}$ & $90.40_{\pm1.24}$ & $88.09_{\pm1.92}$ \\
		& Art             & $70.79_{\pm0.88}$& $72.75_{\pm5.50}$ & $76.53_{\pm1.08}$ & $76.84_{\pm1.02}$ \\
		& Cartoon         & $70.39_{\pm3.21}$& $66.87_{\pm3.46}$ & $75.78_{\pm1.54}$ & $74.05_{\pm5.25}$ \\
		& Sketch          & $70.19_{\pm5.00}$& $74.63_{\pm3.98}$ & $71.43_{\pm3.60}$ & $76.79_{\pm3.38}$ \\
		& Average         & $73.51_{\pm2.19}$& $76.35_{\pm3.41}$ & $78.54_{\pm1.44}$ & $\mathbf{78.94_{\pm1.49}}$
		\\
		\hline\hline
		\multirow{5}{*}{\begin{turn}{90}OfficeHome\end{turn}}
		& Art             & $38.64_{\pm3.14}$ & $39.37_{\pm5.09}$ & $43.52_{\pm0.82}$          & $42.47_{\pm0.66}$ \\
		& Clipart         & $39.28_{\pm4.05}$ & $41.69_{\pm3.32}$ & $41.76_{\pm0.90}$          & $40.58_{\pm1.94}$ \\
		& Product         & $58.73_{\pm1.48}$ & $58.10_{\pm2.36}$ & $59.41_{\pm0.62}$          & $58.00_{\pm1.15}$ \\
		& Real World      & $56.88_{\pm2.22}$ & $60.87_{\pm0.65}$ & $63.91_{\pm0.98}$          & $61.37_{\pm1.47}$ \\
		& Average         & $48.38_{\pm0.51}$ & $50.00_{\pm0.50}$ & $\mathbf{52.15_{\pm0.59}}$ & $50.61_{\pm1.23}$
		\\
		\hline\hline
		\multirow{5}{*}{\begin{turn}{90}VLCS\end{turn}}
		& Caltech101      & $43.37_{\pm32.44}$ &$ 74.08_{\pm12.60}$ & $36.42_{\pm1.10}$ & $85.68_{\pm3.65}$ \\
		& LabelMe         & $52.78_{\pm1.91}$ &$ 59.23_{\pm6.71}$ & $51.49_{\pm0.72}$ & $61.09_{\pm4.98}$ \\
		& SUN09           & $49.88_{\pm1.61}$ &$ 42.96_{\pm6.19}$ & $62.60_{\pm3.90}$ & $50.41_{\pm5.93}$ \\
		& VOC2007         & $27.26_{\pm1.98}$ &$ 41.02_{\pm12.00}$ & $41.87_{\pm4.70}$ & $53.68_{\pm7.61}$ \\
		& Average         & $43.32_{\pm9.13}$ &$ 54.33_{\pm5.14}$ & $48.10_{\pm1.81}$ & $\mathbf{62.72_{\pm3.66}}$
		\\
		\hline\hline
		\multirow{5}{*}{\begin{turn}{90}Terra\end{turn}}
		& Location 38      & $15.00_{\pm13.52}$   & $22.14_{\pm7.50}$ & $28.59_{\pm7.10}$ & $32.32_{\pm18.06}$ \\
		& Location 43      & $14.07_{\pm2.46}$    & $14.07_{\pm1.55}$ & $17.88_{\pm7.10}$ & $25.82_{\pm5.94}$ \\
		& Location 46      & $19.04_{\pm3.18}$    & $21.15_{\pm4.51}$ & $21.77_{\pm3.18}$ & $24.22_{\pm3.59}$ \\
		& Location 100     & $22.14_{\pm16.59}$   & $25.23_{\pm1.99}$ & $40.97_{\pm4.81}$ & $38.38_{\pm9.65}$ \\
		& Average          & $17.56_{\pm2.24}$    & $20.07_{\pm2.92}$ & $27.30_{\pm3.38}$ & $\mathbf{30.19_{\pm4.78}}$
		\\
		\hline
		\end{tabular}}
	\label{tab:results_std}
\end{table}

\begin{table}[htbp!]
	\centering
	\caption{Comparison of FixMatch, StyleMatch and UPLM on labelled seed examples from \cite{zhou2021semi}.}

\scalebox{0.825}{\begin{tabular}{|l|c|c|c|c|}
		\hline
		\multicolumn{1}{|l|}{Target} &
		\multicolumn{1}{c|}{Baseline (FixMatch)} &
		\multicolumn{1}{c|}{StyleMatch} &
		\multicolumn{1}{c|}{UPLM} \\ \hline\hline
        Photo         & $89.18_{\pm0.30}$ & $78.20_{\pm11.30}$ & $\mathbf{91.82_{\pm1.15}}$ \\
        Art           & $73.85_{\pm3.77}$ & $78.10_{\pm1.31}$  & $\mathbf{79.05_{\pm1.74}}$ \\
        Cartoon       & $74.73_{\pm3.72}$ & $\mathbf{82.02_{\pm1.11}}$  & $78.37_{\pm2.08}$ \\
        Sketch        & $74.74_{\pm5.65}$ & $78.60_{\pm1.87}$  & $\mathbf{79.06_{\pm0.52}}$ \\
        Avg.          & $78.12_{\pm1.35}$ & $76.60_{\pm2.77}$  & $\mathbf{82.02_{\pm1.11}}$
		\\
		\hline
  
		\end{tabular}}
	\label{tab:ssdg}
\end{table}
Furthermore, we conducted a thorough comparative analysis, using the labelled seed examples of StyleMatch \cite{zhou2021semi} (Table~\ref{tab:ssdg}). 
 Considering factors like unavailable source code (\cite{yuan2022label}) and relatively large batch sizes(StyleMatch) the comparison of SSDG methods becomes difficult.
We optimized our model by adjusting the batch size to 24 and using the ResNet-50 backbone instead of ResNet-18 \cite{he2016deep}. These modifications were essential for enhancing both performance and computational efficiency. Notably, in comparison with StyleMatch, our method demonstrated superior performance, particularly in the photo domain, using our randomly chosen seeds (available on our GitHub project page) providing a practical and accessible alternative to the examples employed by StyleMatch. 
\subsection{Ablation Study and Analysis}
\label{subsection:ablation_study_analysis}
\begin{figure}[!ht]
  \vspace{-0.2cm}
  \centering
 {\epsfig{file = 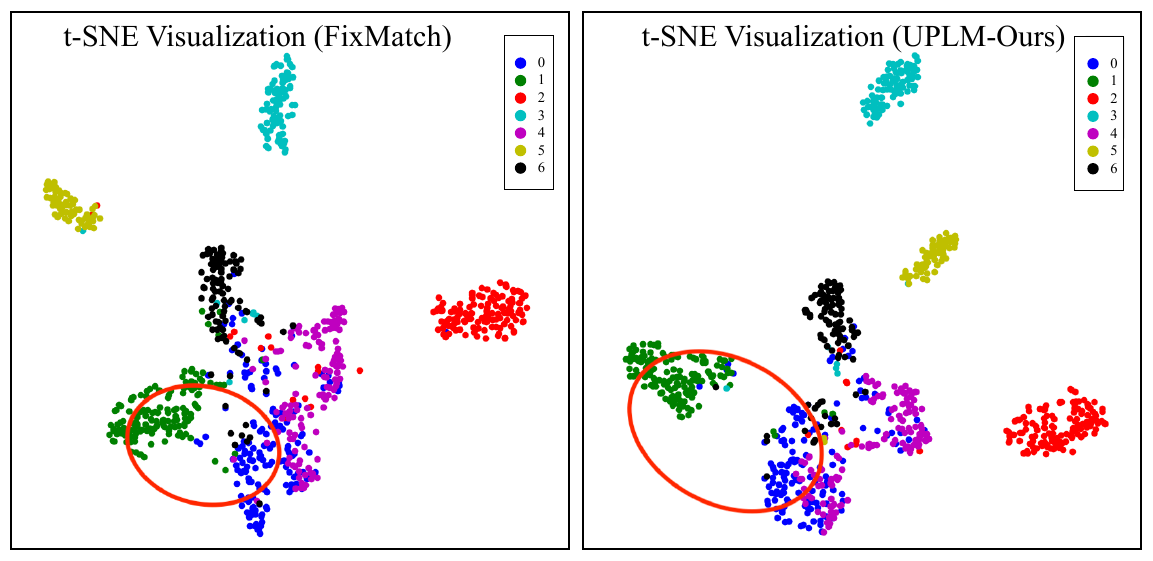, width = 7.5cm}}
\caption{Class-wise feature visualization using t-SNE.}
\label{fig:TSNE}
\end{figure}

\noindent \textbf{t-SNE plots for class-wise features.}
Figure~\ref{fig:TSNE} plots class-wise feature representations obtained using t-SNE for both the FixMatch and UPLM. Our approach facilitates the learning of more discriminative features, resulting in more tightly clustered features within the same class while maintaining greater distance between features belonging to different classes.
\noindent \textbf{Pseudo-labelling accuracy UPL vs FixMatch.}
We compare the accuracy of PLs on the target domains of the PACS dataset (Table~\ref{tab:accuracy}). Results indicate that UPL generates more accurate PLs compared to FixMatch.
\begin{table}[htbp!]
	\centering
	\caption{Comparison of Pseudo-Labelling accuracy (\%).}
	\scalebox{0.7}{\begin{tabular}{|cl|c|c|}
		\hline
		
		&
		\multicolumn{1}{l|}{Target} &
		\multicolumn{1}{c|}{FixMatch} &
		\multicolumn{1}{c|}{UPL} \\ \hline\hline
		\multirow{5}{*}{\begin{turn}{90}PLs Acc. \end{turn}}
		& Photo           & 87.09             & 88.05              \\
		& Art             & 78.80             & 95.93              \\
		& Cartoon         & 83.93             & 89.52              \\
		& Sketch          & 91.55             & 95.30 
		\\
		& Average         & 85.34             & \textbf{92.20}
		\\
		\hline
		\end{tabular}}
	\label{tab:accuracy}
\end{table}
\noindent \textbf{Performance of individual components of MA.} 
Table~\ref{tab:MA ablation} compares the results of six different variants of the model, with each variant utilizing a different strategy for combining the model's parameters during training.  Combining all three models $(\theta_{avg})$, as per our proposal, provides the best performance.
\begin{table}[htbp!]
	\centering
	\caption{Our proposed $\theta_{avg}$ outperforms other variants. }
	\scalebox{0.57}{\begin{tabular}{|cl|c|c|c|c|c|c|c|}
		\hline
		&
		\multicolumn{1}{l|}{Target} &
		\multicolumn{1}{c|}{{$ \theta_{last}$}} &
		\multicolumn{1}{c|}{{$ \theta_{best}$}} &
		\multicolumn{1}{c|}{{$ \theta_{ema}$}} &
		\multicolumn{1}{c|}{{$\theta_{(last+ema)}$}} &
		\multicolumn{1}{c|}{{$\theta_{(last+best)}$}}&
		\multicolumn{1}{c|}{{$\theta_{(best+ema)}$}} &
		{$\theta_{avg}$} \\ \hline\hline
		\multirow{5}{*}{\begin{turn}{90}MA (PACS) \end{turn}}
		& Photo &
		$87.64 $&
        $87.72$ &
		$82.67$ &
		$89.90$ &
		$89.30 $&
		$89.08 $&
		$\mathbf{90.40}$ \\ 
		& Art &
		$72.98 $&
        $73.93$ &
		$70.79 $&
		$\mathbf{78.24}$ &
		$73.11 $&
		$72.38 $&
		$76.53 $\\ 
		& Cartoon &
		$73.93 $&
        $69.16$ &
		$70.39$ &
		$75.43$ &
		$\mathbf{75.90}$ &
		$73.08$ &
		$75.78 $\\ 
		& Sketch &
		$68.72$ &
        $72.23$ &
		$70.19$ &
		$65.48 $&
		$\mathbf{75.31}$ &
		$68.90 $&
		$71.43$ \\ 
		& Average &
		75.82 &
        75.76 &
		73.51 &
		77.26 &
		78.41 &
		75.86 &
		\textbf{78.54} \\ \hline
		\end{tabular}}
	\label{tab:MA ablation}
\end{table}
\noindent \textbf{Performance under various domain shifts.}
We report the performance in various domain shifts in Table~\ref{tab:shift types}, e.g., changes in backgrounds, corruptions, textures, and styles. For instance, background shifts only affect the background of an image and not the foreground object's pixel, texture, and structure \cite{zhang2022delving}. On the other hand, style shifts involve variations in texture, and object parts across different concepts. To evaluate this, we categorize four DG datasets i.e., PACS, VLCS, OfficeHome, and TerraIncognita based on their exhibited shift(s) into the four categories and report results. UPLM outperforms all other methods in all domain shifts, except for a slight advantage of MA in style. The OfficeHome dataset has a limited number of examples per class, leading to potential overfitting due to uncertainty constraints. However, our MA approach demonstrates strong performance by effectively mitigating mis-calibrated PLs.
\begin{table}[htbp!]
	\centering
	\caption{Accuracy (\%) for different types of domain shifts.}
	\scalebox{0.6}{\begin{tabular}{|l|c|c|c|c|c|}
		\hline
		\multicolumn{1}{|c|}{} &
		\multicolumn{1}{|c|}{Texture Shifts} &
		\multicolumn{1}{c|}{Corruption Shifts} &
		\multicolumn{1}{c|}{Background Shifts} &
		\multicolumn{1}{c|}{Style Shifts} \\ \hline
		\multicolumn{1}{|l|}{Methods} &
		\multicolumn{1}{|c|}{PACS} &
		\multicolumn{1}{c|}{Terra} &
		\multicolumn{1}{c|}{VLCS, Terra} &
		\multicolumn{1}{c|}{OH, PACS} \\ \hline\hline
		FixMatch & 73.51 & 17.56 & 30.44 & 60.94 \\
		UPL & 76.35 & 20.07 & 37.20 & 63.17 \\
		MA & 78.54 & 27.30 & 37.70 & \textbf{65.34} \\
		UPLM (Ours) & \textbf{78.94} & \textbf{30.19} & \textbf{46.46} & 64.78 \\
		\hline
		\end{tabular}}
	\label{tab:shift types}
\end{table}

\section{CONCLUSION}
\label{sec:Conclusion}
We presented a new SSDG approach (UPLM) featuring uncertainty-guided pseudo-labelling and model averaging mechanisms. The proposed approach leverages the model's predictive uncertainty to develop a pseudo-labelling selection criterion that mitigates the impact of poor model calibration under multi-source unlabelled data. The model averaging technique reduces overfitting to source domains in the presence of limited labels and domain shifts. Results on several challenging DG datasets suggest that our method provides notable gains over the baseline. We believe that our work will encourage the development of more data-efficient visual recognition models that are also generalizable across different domains.

\section*{\uppercase{Acknowledgements}}
\label{acknowledgments}
This work was supported in part by Google unrestricted gift 2023. The authors are grateful for their generous support, which made this research possible.
\bibliographystyle{apalike}
{\small
\bibliography{review}}
\section*{APPENDIX}
\renewcommand{\thesubsection}{\Alph{subsection}}
\subsection{Uncertainty $\kappa$ vs ECE plot}
\label{appendix:ece}
For each training iteration, we first calculate the mean of uncertainty ($\mathcal{V}$) for all input examples in a batch across class dimension to obtain the overall uncertainty for each example. Also, we compute the corresponding ECE score for this batch. Next, after each epoch, we compute mean of overall uncertainty (for each example) over all examples seen and also compute the mean over ECE. Each epoch yields a pair of mean uncertainty and corresponding mean ECE over all examples. We then sort the mean uncertainty values in ascending order to build x-axis and the corresponding mean (ECE) to plot y-axis.
\subsection{Analysis of Hyperparameters}
\label{appendix:hyperparameters} 
 \begin{table}[H]
	\centering
	\caption{Ablation of $\kappa$ in range [0.2, 0.8]}
	\scalebox{0.6}{\begin{tabular}{|cl|c|c|c|c|c|c|c|}
		\hline
		&
		\multicolumn{1}{l|}{Target} &
		\multicolumn{1}{c|}{0.2} &
		\multicolumn{1}{c|}{0.3} &
		\multicolumn{1}{c|}{0.4} &
		\multicolumn{1}{c|}{0.5} &
		\multicolumn{1}{c|}{0.6} &
		\multicolumn{1}{c|}{0.7} &
		\multicolumn{1}{c|}{0.8} \\ \hline\hline
		\multirow{5}{*}{\begin{turn}{90}PACS\end{turn}}
		& Photo           & 87.07 & 86.11 & 86.77 & 78.32 & 84.79 & 70.42 & 65.09 \\
		& Art             & 75.73 & 71.58 & 68.70 & 69.73 & 68.02 & 57.13 & 60.40 \\
		& Cartoon         & 68.09 & 66.3  & 65.02 & 61.56 & 63.14 & 59.68 & 59.90 \\
		& Sketch          & 79.38 & 72.13 & 73.12 & 59.96 & 71.21 & 46.17 & 66.45 \\
		& Average         & \textbf{77.57} & 74.03 & 73.40 & 67.39 & 71.79 & 58.35 & 62.96\\
		\hline\hline
		\multirow{5}{*}{\begin{turn}{90}OfficeHome\end{turn}}
		& Art             & 42.89 & 37.99 & 42.69 & 43.22 & 38.90 & 42.89 & 40.34  \\
		& Clipart         & 40.92 & 42.15 & 37.55 & 42.50 & 39.31 & 36.70 & 41.05  \\
		& Product         & 57.92 & 58.95 & 56.25 & 58.71 & 56.32 & 57.02 & 57.02  \\
		& Real World      & 63.76 & 59.44 & 60.80 & 62.98 & 61.65 & 58.30 & 59.93  \\
		& Average         & 51.37 & 49.63 & 49.32 & \textbf{51.85} & 49.05 & 48.73 & 49.59  \\
		\hline\hline 
		\multirow{5}{*}{\begin{turn}{90}VLCS\end{turn}}
		& Caltech101      &29.40	&  57.03 & 50.46  & 85.72 & 80.71 & 87.42 & 80.42   \\
		& LabelMe         &54.14	&  53.77 & 55.76 & 60.69 & 54.18 & 64.72 & 60.47   \\
		& SUN09           &65.42	&  60.02 & 56.79 & 52.16 & 51.22 & 43.57 & 49.18   \\
		& VOC2007         &33.56	&  48.13 & 35.55 & 36.58 & 38.00 & 51.18 & 39.19   \\
		& Average         &45.63	&  54.74 & 49.64 & 58.79 & 56.03 & \textbf{61.72} & 57.32   \\
		\hline\hline
		\multirow{5}{*}{\begin{turn}{90}Terra\end{turn}}
		& Location 38      & 22.09   & 12.26 & 4.18  &	39.54 &	44.48 &	11.54 &	33.57      \\
		& Location 43      & 10.96	 & 22.80 & 15.19 &	25.84 &	16.73 &	20.10 &	12.95  \\
		& Location 46      & 23.97	 & 19.46 & 23.98 &	20.57 &	16.66 &	22.49 &	22.11  \\
		& Location 100     & 39.72	 & 48.20 & 32.40 &	37.63 &	41.95 &	30.65 &	32.38  \\
		& Average          & 24.19	 & 25.68 & 18.94 &	\textbf{30.90} &	29.96 &	21.20 &	25.25  \\
		\hline
		\end{tabular}}
	\label{tab:supp-ablation}
\end{table}

 \subsubsection{Computational Cost of MC Forward Passes}
We use $\mathcal{N}$ = 10 Monte Carlo (MC) forward passes in all experiments, with negligible computational overhead. The per-iteration execution times in milliseconds for different values of $\mathcal{N}$ (1, 5, 10, 20, 40, 80, 160) are 134.6, 135.1, 135.6, 137.5, 138.5, 141.7, 146.6, respectively.
 
 \subsubsection{Accuracy with Changing the Amount of Unlabelled Data $\mu$}
The average accuracies on PACS for $\mu= [1-6]$ are 65.25, 71.90, 75.47, 73.7, \textbf{78.94}, 78.22, respectively. 
$\mu$ = 5 performs best overall, which is used throughout in all our experiments. Note that, $\mu$ values beyond 6 are not possible due to computational constraints.
\subsubsection{Effect of Certainty Threshold}

We present an ablation study concerning the selection of the certainty threshold $\kappa =$ 0.2 (indicating the least certainty) to $\kappa =$ 0.8 (indicating the highest certainty), as detailed in Table~\ref{tab:supp-ablation}. 
\end{document}